\documentclass{article}

 \usepackage[dblblindworkshop, final]{neurips_2025}
\usepackage{graphicx}
\usepackage[utf8]{inputenc} 
\usepackage[T1]{fontenc}    
\usepackage{hyperref}       
\usepackage{url}            
\usepackage{booktabs}       
\usepackage{amsfonts}       
\usepackage{nicefrac}       
\usepackage{microtype}      
\usepackage{footnote}
\makesavenoteenv{table}
\usepackage{amsmath}
\usepackage[table,xcdraw]{xcolor}
\title{Catalyst GFlowNet for electrocatalyst design:\\ A hydrogen evolution reaction case study}

\author{Lena Podina\thanks{Work done while a research intern at Mila Quebec AI Institute, Montreal, Canada} \\
	University of Waterloo\\	\texttt{lpodina@uwaterloo.ca} \\
	\And
	{Christina Humer} \\
        ETH Zurich\\
	\AND
        Alexandre Duval \\
	Entalpic \\
        \And
        Victor Schmidt \\
	Entalpic \\
        \And
        Ali Ramlaoui \\
	Entalpic \\
        \And
        Shahana Chatterjee \\
	Mila Quebec AI Institute \\
        \And
        Yoshua Bengio \\
	Mila Quebec AI Institute \\
	\And
	Alex Hernandez-Garcia \\
	Mila Quebec AI Institute \\
	\And
	David Rolnick \\
	Mila Quebec AI Institute \\
        \And
        Félix Therrien \\
	Mila Quebec AI Institute \\
    \texttt{felix.therrien@mila.quebec} \\
}

\begin{document}

\maketitle

\begin{abstract}
Efficient and inexpensive energy storage is essential for accelerating the adoption of renewable energy and ensuring a stable supply, despite fluctuations in sources such as wind and solar. Electrocatalysts play a key role in hydrogen energy storage (HES), allowing the energy to be stored as hydrogen. However, the development of affordable and high-performance catalysts for this process remains a significant challenge. We introduce Catalyst GFlowNet, a generative model that leverages machine learning-based predictors of formation and adsorption energy to design crystal surfaces that act as efficient catalysts. We demonstrate the performance of the model through a proof-of-concept application to the hydrogen evolution reaction, a key reaction in HES, for which we successfully identified platinum as the most efficient known catalyst. In future work, we aim to extend this approach to the oxygen evolution reaction, where current optimal catalysts are expensive metal oxides, and open the search space to discover new materials. This generative modeling framework offers a promising pathway for accelerating the search for novel and efficient catalysts.
\end{abstract}

\section{Introduction}


As society moves away from fossil fuels and towards renewable energy sources, technological advances such as renewable energy storage and efficient energy production become key drivers in this transition~\cite{oc_overview,freeman2024impact}. Since peak energy usage hours differ from energy production hours~\cite{pitra2021duck}, known as the ``duck'' curve, energy storage can ensure that cities obtain the energy they need at the right time. However, methods suitable for large-scale storage, such as hydrogen energy storage (HES)~\cite{HES}, require effective catalyst materials both for energy storage (via electrolysis) and for energy retrieval (via fuel cells)~\cite{oc_overview,osman2022hydrogen}. Most catalysts that are currently relied upon to accelerate both energy storage and retrieval are expensive to manufacture~\cite{lim2022non}. For this reason, the search for efficient, cheap, and durable catalysts is vital to increasing the adoption of renewable energy.
    
Machine learning, growing in its applications and reaching state-of-the-art performance in image generation~\cite{croitoru2023diffusion}, language modelling~\cite{vaswani2017attention}, and scientific discovery~\cite{wang2023scientific}, has been applied to predict catalyst properties and generate new catalysts~\cite{catalyst_vae, faenet,oc20,oc22}. Graph neural networks such as FAENet~\cite{faenet,schnet,gemnet} are able to predict adsorption energies, which are a key predictor of catalytic efficiency~\cite{Norskov_HER,Norskov_OER}. Methods such as Crystal-GFN~\cite{crystal_gfn} are capable of generating molecules and crystal structures within specified symmetry constraints, which meet certain desirable properties. These machine learning methods can be broadly divided into two categories, predictive models (used for catalyst evaluation)~\cite{zitnick2020introduction, duval2023hitchhiker}, and generative models (used for catalyst design)~\cite{catalyst_vae,catgpt}.

\begin{figure}
    \centering
    \includegraphics[width=0.7\linewidth]{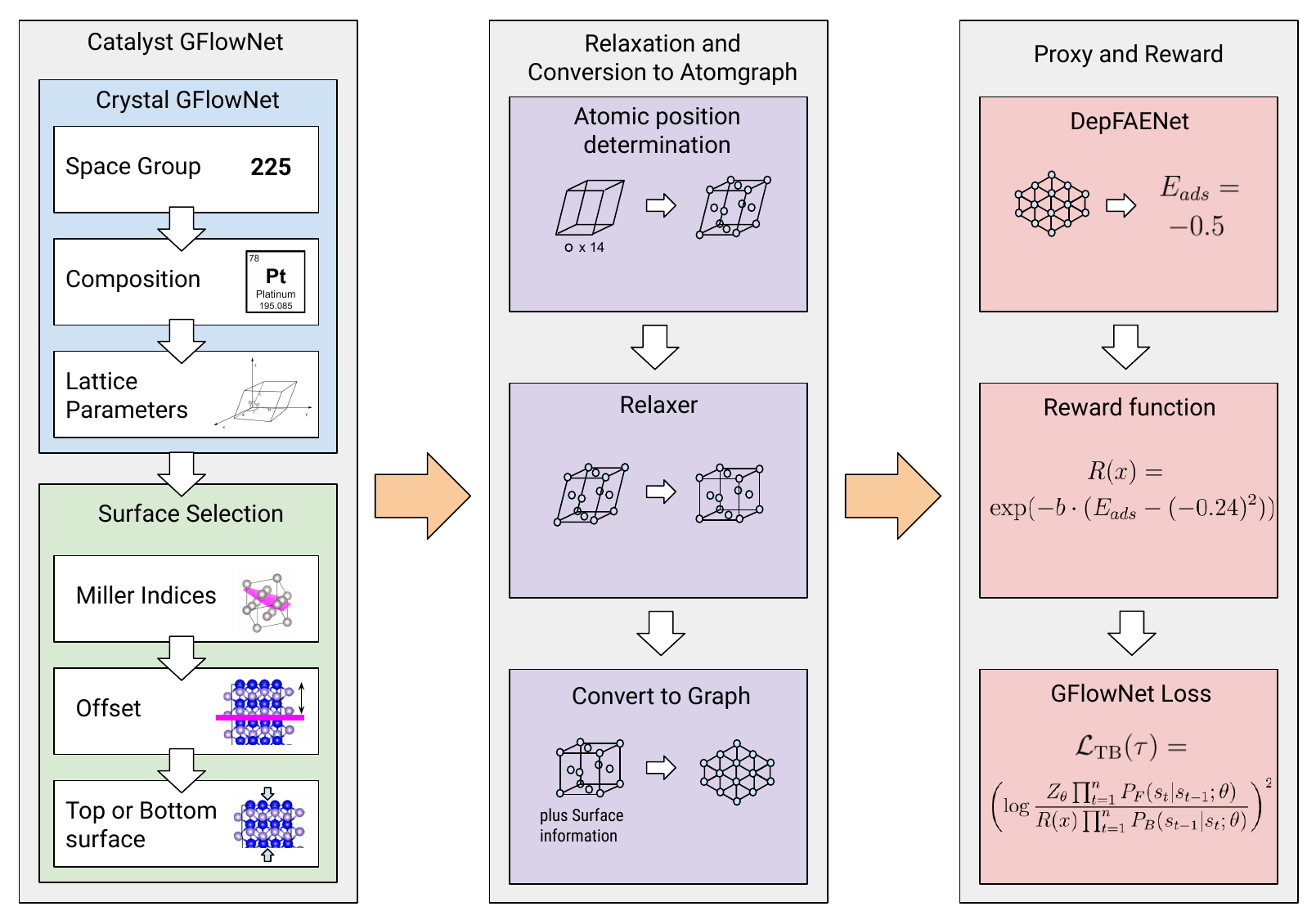}
    \caption{Overview of the Catalyst GFlowNet framework. The leftmost section, Catalyst GFlowNet, samples the catalyst surface. The middle section determines atom positions, relaxes the structure, and converts to a graph. The rightmost section obtains the adsorption energy from a predictive model and the reward function to train the GFlowNet.}
    \label{fig:catalyst_gfn}
\end{figure}

Here, we introduce Catalyst GFlowNet (Figure~\ref{fig:catalyst_gfn}), a novel catalyst discovery method building on Crystal-GFN~\cite{crystal_gfn}, designed to generate a diverse set of efficient catalyst materials. Generative flow networks (GFlowNets or GFNs) are generative models that are able to sample sequentially constructed objects from a desired distribution. This property allows our framework to propose a variety of catalyst materials rather than a single optimal material. Subsequent real-world experiments have a greater chance of discovering a truly high-quality catalyst material when they are able to test multiple proposed materials. Our approach for catalyst design builds on crystal generation by constructing a periodic crystal and then cutting a surface that acts as the catalyst. Additionally, we integrate ML-based structure relaxations into our framework to ensure that the final samples are relaxed (stable). As a signal on the quality of the catalyst material, we use a Graph Neural Network (GNN) based on FAENet~\cite{faenet} to estimate the adsorption energy for a given adsorbate molecule in a reaction of interest. A tailored reward function ensures that we generate realistic crystals that can then serve as the basis for efficient catalysts. In our case study, we verify that Catalyst GFlowNet is able to rediscover the best-known catalysts, demonstrating its promising value for catalyst design.

\section{Methodology}\label{sec:methodology}

In this section, we introduce the general architecture and workflow of the Catalyst GFlowNet (Section~\ref{sec:architecture}) and the experimental setup for the hydrogen evolution reaction (HER) case study (Section~\ref{sec:HER}). Generally, a GFlowNet learns to sample objects proportionally to their reward; we leverage this property while imposing constraints on the way that samples are constructed. Background material on GFlowNets and catalysts for HER is covered in the appendix, Section~\ref{sec:background}.
\subsection{Architecture}\label{sec:architecture}

The Catalyst GFlowNet builds on the Crystal-GFN framework. The overall framework, depicted in Figure~\ref{fig:catalyst_gfn}, consists of three sequential steps: sample generation, sample preparation, and reward computation. After the reward is computed, a GFlowNet objective, such as the trajectory balance loss~\cite{trajectory_balance}, can be used to train the GFlowNet to sample catalyst representations proportionally to the reward function.

In the first phase, the catalyst surface is sequentially constructed by the GFlowNet. Here, we Crystal-GFN to build catalyst surfaces given a crystal structure. In particular, this involves first selecting the space group (integer between 1 and 230, possibly decomposed into lattice system and point group); following this, selecting the composition (e.g. $\text{A}_\text{x}\text{B}_\text{y}\text{C}_\text{z}$); and finally, selecting the lattice parameters (lattice vectors $a,b,c$ and angles between them $\alpha, \beta, \gamma$). Conditioned on this crystal representation, the Catalyst GFlowNet selects the Miller indices~\cite{wandelt2018encyclopedia} (three integers ranging from -2 to 2) representing the surface of the periodic crystal object that would act as the catalyst. For example, the Miller indices (100) cut a plane that is parallel to the Y-Z plane created by the lattice vectors\footnote{Miller indices $hkl$ denote the 3D plane (and those parallel to it) that intersects the $x$-axis at 1/$h$, the $y$-axis at 1/$k$, the $z$-axis at 1/$l$.}. Following this, the offset for the surface plane is selected (continuous value between 0 and 1), which represents the location of the cut in the direction perpendicular to the Miller plane. Finally, the network selects whether the top or bottom surface (represented by a boolean) is used. For a single-element material, such as platinum, the offset and top/bottom surface boolean do not change the resulting surface. However, for more complex materials, such as oxides, the offset and the surface boolean can lead to surfaces with widely different properties. 

In the second phase (middle part of Figure~\ref{fig:catalyst_gfn}), we obtain atomic positions symmetrically compatible with the crystal representation by using PyXtal's structure generator~\cite{pyxtal}. 
After the structure is fully determined, M3GNet~\cite{m3gnet} is used to relax the atoms and the unit cell. Then, a converter module creates a surface object using the final structure and sampled Miller indices, top surface information, and offset. Finally, this surface object is converted into a graph, where each node represents an atom. The adsorbate is its own graph, disconnected from the surface~\cite{carbonero2023importance}. 

In the final phase (right-most side of Figure~\ref{fig:catalyst_gfn}), the atom graph is passed to the proxy model, DepFAENet in our case, which is a GNN based on FAENet~\cite{faenet}, equivariant to Euclidean symmetries, trained to predict adsorption energies for different adsorbates. DepFAENet differs from the original FAENet implementation in that the surface and adsorbate are disconnected graphs~\cite{carbonero2023importance}. The model predicts the adsorption energy directly without situating the adsorbate on the surface.
Then, the predicted adsorption energy is passed to the reward function (described in detail in Section~\ref{sec:HER}). The rewards obtained for the constructed samples are then used to compute the trajectory balance loss to train the GFlowNet.


        

\subsection{Case study: Hydrogen Evolution Reaction}\label{sec:HER}

For this case study, our aim is to generate stable catalysts for the hydrogen evolution reaction (HER) that have the smallest possible \textit{overpotential}. In catalysis, the overpotential ($\eta$) is the amount of electric potential that needs to be applied on top of the theoretical reaction potential; it is a measure of the efficiency of a chemical reaction and its catalyst. For HER, in acidic conditions, it can be predicted directly from the adsorption free energy of hydrogen~\cite{Norskov_HER} (which is what our GNN proxy is trained to predict) plus a small correction. Here, to obtain proposed catalysts ($x$) with low overpotentials, we use the following reward function:

\begin{equation} \label{eq:reward}
R(x) = \exp(-b \eta^2) \quad \text{where} \quad \eta = E_{H}(x) + E_{corr}
\end{equation}


We set $b=100$ and the Gibbs correction $E_{corr}=-0.24$. Although the reward only takes the overpotential ($\eta$) into account, structure relaxation for each sample (middle component of Figure~\ref{fig:catalyst_gfn}) ensures that we generate reasonably stable structures. Additionally, at sampling time, structures that have high formation energy among structures with the same composition are discarded.



As a proof of concept, we test our workflow on a highly restricted search space, where atomic positions are fully determined given the space group and the number of atoms. In this search space, we have an expectation of what materials should be sampled the most. With other methods, such as those based on large language models~\cite{catgpt}, imposing constraints on crystal symmetry, space group, and number of atoms would be much more difficult, if not impossible. The search space that we consider matches closely the catalysts described in Nørskov et al.~\cite{Norskov_HER}. The full description of the experimental setting can be found in Table~\ref{tab:HER_setup} in the appendix. We consider 12 possible elements, including platinum, the best known HER catalyst~\cite{Norskov_HER}. We consider the space of single-element structures only, as per Nørskov et al.~\cite{Norskov_HER}. Taking the most common form of each structure as found in the Materials Project~\cite{materials_project}, we note the space group of each structure and restrict the search space to those space groups only. These are space groups $Fm\overline{3}m$ (225) and $Im\overline{3}m$ (229). 

\section{Results}

In this section, we analyze the catalysts sampled by the Catalyst GFlowNet after being trained, following the experimental setup for the hydrogen evolution reaction (HER). We sampled 1000 structures and performed a M3GNet-based relaxation on the crystal lattice. Then, we filter out all structures that relax to a high formation energy.  Specifically, we compute the minimal sampled formation energy per composition, and keep only those relaxed structures that have a formation energy within 0.05 eV of the minimum energy. We do this because in some cases the relaxation strays too far from the true basin given the starting lattice parameters, and as such would not be a stable structure. Then, for each composition, we keep only the space group that has the lowest minimum energy, since the one with the higher energy is estimated to be unstable. Finally, we cut the surface according to the parameters chosen by the Catalyst GFlowNet to construct the final catalyst surface.

\begin{figure}
    \centering
    \includegraphics[width=1.0\linewidth]{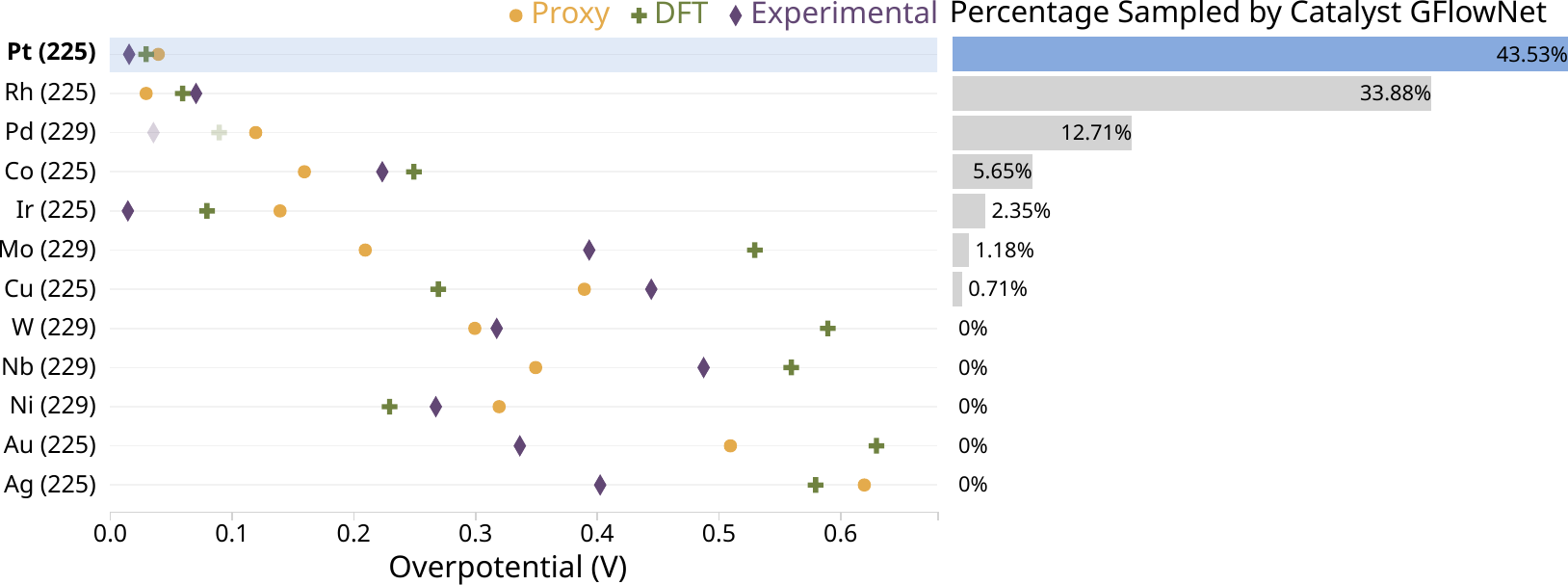}
    \caption{Proportion of sampled structures (right) for the hydrogen evolution reaction case study. Low overpotential (left) is a predictor for efficient catalysts and aligns with the higher sampling rates. The values are detailed in Table~\ref{tab:HER_results} in the appendix. Pd(229) does not have experimental or DFT overpotentials, corresponding values for Pd(225) are displayed with shaded colors for reference.}
    \label{fig:HER_results}
\end{figure}

The filtered set of relaxed structures contains 425 structures. Figure~\ref{fig:HER_results} details the percentage of samples of each of the structures in the paper. As expected, the most frequently sampled structures are those that have a low overpotential, as predicted by the proxy model and DFT values by Nørskov et al.~\cite{Norskov_HER} (see Figure~\ref{fig:HER_results}, left). Platinum (Pt) is known to be the best catalyst for HER, but rhodium (Rh) also performs well, as evidenced by experimental, DFT, and proxy values. We find in fact that rhodium and platinum are the most sampled structures. All generated structures except palladium with space group 229 are found in Nørskov et al.~\cite{Norskov_HER} (and are experimentally observed~\cite{materials_project}). Palladium with space group 225, the one that is experimentally observed, was sampled, but filtered out because its M3GNet-predicted formation energy was slightly higher than its 229 counterpart. Overall, this experiment validates that the Catalyst GFlowNet framework can independently rediscover the best known HER catalysts, and sample them proportionally to their performance.

\section{Discussions and Conclusion}

In this paper, we propose a novel catalyst discovery framework that generates complete and relaxed crystal structure descriptions (including atomic positions) for the purpose of finding an inexpensive, efficient catalyst for renewable energy storage. We show that the method can find the best currently known catalyst for the hydrogen evolution reaction (i.e., platinum). We also show that our method samples diverse structures, increasing the likelihood of finding suitable catalysts that withstand real-world testing. Finally, we generate sample structures that our model considers to be good catalysts for the hydrogen evolution reaction; further investigation could shed light on whether these structures could perform well experimentally. This framework is the prototype of a novel method to generate \emph{de novo} stable catalysts for a variety of materials science tasks; integration with real-world experiment and active learning approaches would surely have an impact in the search to reduce the cost of renewable energy storage. In future work, we aim to extend our method in several ways. We aim to test the framework on the oxygen evolution reaction, for which there is currently no efficient, inexpensive catalyst. 
Additionally, we aim to add functionality to the Catalyst GFlowNet in order to sample atomic positions. We foresee the extension of this work by embedding the GFlowNet in a multi-fidelity active learning loop \cite{hg2024mfgfn}.

\bibliographystyle{plain}
\bibliography{references}  

\newpage
\appendix

\section{Background}\label{sec:background}

In this section, we describe Generative Flow Networks in detail (Section~\ref{sec:gflownets}), and we perform a survey of related work that tackles the catalyst discovery problem (Section~\ref{sec:related_work}).

\subsection{Catalysts for hydrogen energy storage}\label{sec:crystals_and_catalysts} 


Hydrogen Energy Storage (HES) is a scalable, sustainable option for storing excess energy for later use. Although HES has a relatively low efficiency compared to alternatives such as pumped storage hydroelectricity and battery storage~\cite{oc_overview,osman2022hydrogen,green_hydrogen}, it is more scalable and transportable than the alternatives. HES via water electrolysis (water-splitting) can be thought of as two half-cell reactions: the hydrogen evolution reaction and the oxygen evolution reaction. Both HER and OER are electrochemical reactions which occur in the catalyzer at different electrodes. While HER is a one-step reaction, OER consists of four intermediate reactions which, under an applied voltage, produces oxygen gas, protons and electrons under acidic conditions. HER, in this setting, brings electrons and protons together to produce hydrogen gas. As the energy is used for water electrolysis, the hydrogen gas can be collected and stored in tanks, or other storage containers. When the energy is in demand again, fuel cells can extract the energy from the stored hydrogen at about 60\% efficiency~\cite{fan2021recent}. An efficient catalyst, used in the fuel cell or in the electrocatalyzer, can significantly speed up the reaction rate and improve the efficiency of the process~\cite{oc_overview}.




\subsection{Generative Flow Networks}\label{sec:gflownets}

A generative flow network (GFlowNets, GFN) is an inference method introduced by~\cite{bengio2021flow} that allows one to sequentially construct objects with probability proportional to a non-negative reward function. They have been employed in tasks such as molecule generation, crystal generation, and for causal discovery~\cite{RGFN,crystal_gfn,DynGFN}. A desirable capability of GFlowNets is their ability to generate a diversity of objects that all have high reward (relative to other possible candidates in the space). Within catalyst discovery, this is key, as the reward function may not capture all the relevant predictors of whether a material will be a good catalyst. For example, the reward function may capture a target adsorption energy for the catalyst, but not its stability, which is an important trait (high stability) if the catalyst is to be used in real-world applications. After generating a variety of high-reward catalyst materials,  one or more of these materials may also be found to be stable during experimental testing. 
Furthermore, GFlowNets perform particularly well in cases where the search space is quite large, and outperform methods such as MCMC and reinforcement learning in terms of mode mixing~\cite{gflownet_foundations}. 


A key component of GFlownets is the proxy model, which is how the GFlowNet obtains information on the reward landscape, and hence what ``desirable'', high-reward samples are. Proxy models in our context are predictive models (possibly machine learning models) that allow one to evaluate the reward of a particular sampled object. In the case of molecule or crystal discovery, it may be that the sampled molecule or crystal does not exist in any database, and as such, relevant properties of the crystal must be computed in real time instead. Here, we note that for certain properties such as the formation energy of a crystal, physics simulations such as density functional theory (DFT) are traditionally performed. These simulations are computationally expensive and not scalable to high batch sizes. Hence, a machine learning model that is able to replace the physics simulation and compute the properties of interest is valuable; for crystals, there are many works that develop machine learning-based approaches to compute properties such as formation energy and adsorption energy~\cite{schnet, painn, faenet, m3gnet}.

Crystal GFlowNet is a GFlowNet designed to discover new materials by sequentially sampling crystal structures~\cite{crystal_gfn}. This work uses the GFlowNet paradigm to construct and sample crystal structures that have a high reward. The authors test the method on a reward based upon formation energy, where a high reward corresponds to a low formation energy, and a low reward corresponds to a high formation energy. This allows the framework to be used to sample thermodynamically stable materials with desirable properties. As mentioned earlier, the reward function may use any measurable (or predicted, via a proxy model) characteristic of the material. Sequential crystal construction proceeds in three steps. First, the space group (symmetry operations) of the crystal are selected. Then, the composition (elements and their ratios), and finally, the lattice parameters, which determine the shape of the final lattice. Atom positions are not sampled by the model.



    

\subsection{Machine learning for catalyst discovery}\label{sec:related_work}


Since traditional catalyst design methods are time-consuming and involve trial and error experiments within a very large search space~\cite{hautier2012computer}, scientists are turning to machine learning approaches to more efficiently explore the space of good catalysts. Recent work has leveraged both predictive models (which evaluate the performance of catalysts based on their measurable attributes) and generative models (which aim to construct catalyst representations with desirable properties) for catalyst discovery. 

One major thread of predictive machine learning for catalyst design are predictive models to replace density functional theory (DFT) calculations of adsorption energies. The Open Catalyst Dataset 2020~\cite{oc20} and 2022~\cite{oc22} are datasets of 1.2 million and 62 thousand relaxations respectively. The datasets span a wide variety of materials, surfaces and adsorbates, with the inclusion of oxides in OC22. Machine learning models such as FAENet, Schnet, PaiNN, and Graphormer~\cite{faenet} are neural networks (e.g. graph neural networks, transformers) that can be trained to perform the OC20 Initial Structure to Relaxed Energy (IS2RE) task, which involves predicting the relaxed energy of an adsorbate-catalyst system from their initial atomic positions. Machine learning models have also been used to predict formation energies, perform relaxations, predict forces~\cite{m3gnet}. Predictive machine learning models have also been used to predict the thermodynamic stability of catalysts~\cite{zou2025predicting}. All of these predictive models have the potential to be included as proxy models within generative approaches.

Recent generative approaches to catalyst discovery include reinforcement learning~\cite{adsorbRL}, large language models~\cite{lai2023artificial}, variational autoencoders~\cite{schilterVAE}, and generative adversarial networks~\cite{ishikawaGAN} which aim to generate catalysts \emph{de novo}, by meeting criteria such as target adsorption energies and encoding this in the reward or loss. AdsorbRL~\cite{adsorbRL} employs Deep Q-learning to find compositions of catalysts that have either minimal or maximal adsorption energy. They only model composition, omitting space groups, crystal symmetries, surface selection, and atom positions.~\cite{schilterVAE} addresses the problem by employing a VAE to generate catalysts by sampling from the latent space, but we note that reliance on the latent space makes controllable generation and systematic exploration of possible catalysts difficult.~\cite{ishikawaGAN} takes the approach of GANs to generate catalysts for the ammonia formation reaction, but does not take catalyst stability into account during generation. LLM-based tools such as~\cite{lai2023artificial,catberta} have so far been mainly applied to help screen known catalysts and perform a literature search.

\newpage
\section{Effect of relaxation on lattice parameters}

\begin{figure}[h]
    \centering
    \includegraphics[width=0.7\linewidth]{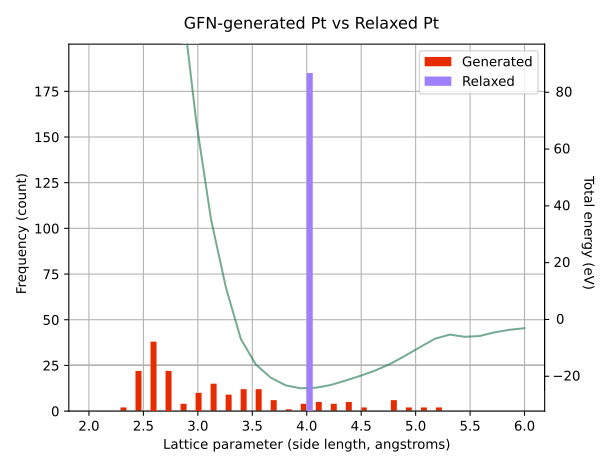}
    \caption{For the HER case study, the platinum samples generated by the GFlownet relax to the minimum energy every time. The green line represents the total energy of the system.}
    \label{fig:relaxing_platinum}
\end{figure}

After the GFlowNet generates samples, we relax them using M3GNet PES~\cite{m3gnet}, as detailed in Section~\ref{sec:architecture}. Figure~\ref{fig:relaxing_platinum} demonstrates the change in lattice parameters of the cubic platinum lattice before and after relaxation. We note that the lattice parameters relax to the minimum of the total energy of the system.

Table~\ref{tab:HER_setup} details the search space for the Hydrogen evolution reaction case study (Section~\ref{sec:HER}). We note that in general, the Catalyst GFlowNet's constraints and search space can be set to be much larger than that used in the HER case study. In future work, we will consider a larger search space up to 80-100 atoms, more elements and space group options, while enforcing a neutral charge constraint.

\begin{table}[h]
  \caption{Hydrogen evolution reaction catalyst search space. }
  \label{tab:HER_setup}
  \centering
  \begin{tabular}{lrl}
    \toprule
    Setting     & Value     & Units \\
    \midrule
    Possible elements     & Pt, Ag, Au, Pd, Ir, Ni, W, Co, Cu, Mo, Rh, Nb     &  \\
    Min. \# of different elements in unit cell     & 1     &  \\
    Max. \# of different elements in unit cell     & 1     &  \\
    Min. atoms in unit cell      & 2     &  \\
    Max. atoms in unit cell     & 4     &  \\
    Min. atoms per element     & 2     &  \\
    Max. atoms per element     & 4     &  \\
    Enforce neutral charge?     & No     &  \\
    Possible space groups     & 225, 229     &  \\
    Min. lattice parameter length     & 2     & Angstroms \\
    Max. lattice parameter length     & 6     & Angstroms \\
    Min. angle between lattice vectors     & 60     & Degrees \\
    Max. angle between lattice vectors      & 140     & Degrees \\
    \bottomrule
  \end{tabular}
\end{table}

\newpage
\section{Table of hydrogen evolution reaction samples}

\begin{table}[h]
  \caption{Proportions of sampled structures for the case study: hydrogen evolution reaction.}
  \label{tab:HER_results}
  \centering
  \begin{tabular}{llrrrrr}
    \toprule
    \multicolumn{2}{c}{} & \multicolumn{3}{c}{Overpotential (eV)} & \multicolumn{2}{c}{Samples}                   \\
    \cmidrule(r){3-5} \cmidrule(r){6-7}
    Composition     & \parbox[t]{1cm}{Space Group}     & Experimental\footnote{When available, the experimental overpotential was added from~\cite{danilovic2012enhancing}. For the remaining elements, marked with an asterisk, the overpotentials were inferred using a linear fit between the log of experimental exchange current density (from~\cite{trasatti1972work}) and the available experimental overpotentials (from~\cite{danilovic2012enhancing}). Figure~\ref{fig:fit_overpotential} shows the linear fit and predictions.} & DFT & \parbox[t]{1cm}{Proxy Model}& Count     & Percentage \\
    \midrule
    \textbf{Pt} & 225& $0.016\hspace{0.5em}$ & $0.03\hspace{0.5em}$ & $0.04$     & 185     & \textbf{43.53} \\
    Rh     & 225     & $0.071^*$ & $0.06\hspace{0.5em}$ & $0.03$     & 144     & 33.88 \\
    Pd     & 229     & $0.036^*$ & $0.09^*$ & $0.12$     & 54     & 12.71 \\
    Co     & 225     & $0.224^*$ & $0.25\hspace{0.5em}$ & $0.16$     & 24     & 5.65 \\
    Ir     & 225     & $0.015\hspace{0.5em}$ & $0.08\hspace{0.5em}$ & $0.14$     & 10     & 2.35 \\
    Mo     & 229     & $0.394^*$ & $0.53\hspace{0.5em}$ & $0.21$     & 5     & 1.18 \\
    Cu     & 225     & $0.445\hspace{0.5em}$ & $0.27\hspace{0.5em}$ & $0.39$     & 3     & 0.71 \\
    W      & 229     & $0.318^*$ & $0.59\hspace{0.5em}$ & $0.30$     & 0     & 0.00 \\
    Nb     & 229     & $0.488^*$ & $0.56\hspace{0.5em}$ & $0.35$     & 0     & 0.00 \\  
    Ni     & 229     & $0.268\hspace{0.5em}$ & $0.23\hspace{0.5em}$ & $0.32$     & 0     & 0.00 \\
    Au     & 225     & $0.337\hspace{0.5em}$ & $0.63\hspace{0.5em}$ & $0.51$     & 0     & 0.00 \\
    Ag     & 225     & $0.403\hspace{0.5em}$ & $0.58\hspace{0.5em}$ & $0.62$     & 0     & 0.00 \\ 
    \midrule
    Sum    &      &  &  &     & 425     & $100.00$ \\
    
    \bottomrule
  \end{tabular}
\end{table}

\begin{figure}[h]
    \centering
    \includegraphics[width=0.8\linewidth]{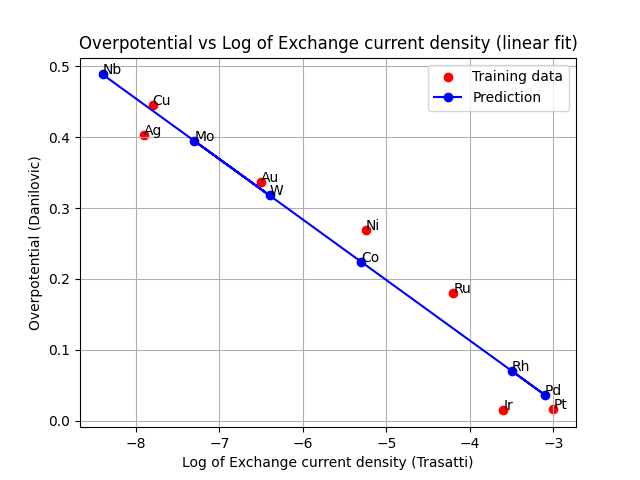}
    \caption{A linear fit that maps log of exchange current density to experimental overpotential using data from~\cite{trasatti1972work} and~\cite{danilovic2012enhancing}. These values are used column 3 (from the left) of Table~\ref{tab:HER_results}.}
    \label{fig:fit_overpotential}
\end{figure}

\end{document}